\documentclass[10pt,twocolumn,letterpaper]{article}

\usepackage{wacv}              

\usepackage{graphicx}
\usepackage{amsmath}
\usepackage{amssymb}
\usepackage{booktabs}
\usepackage[accsupp]{axessibility} 

\usepackage[pagebackref,breaklinks,colorlinks]{hyperref}

\usepackage[capitalize]{cleveref}
\crefname{section}{Sec.}{Secs.}
\Crefname{section}{Section}{Sections}
\Crefname{table}{Table}{Tables}
\crefname{table}{Tab.}{Tabs.}

\def\wacvPaperID{527} 
\def\confName{WACV}
\def\confYear{2025}

\usepackage{graphicx}
\usepackage{floatrow}
\usepackage{algorithm}
\usepackage{algorithmic}

\usepackage{hhline}
\usepackage{tablefootnote}
\usepackage[section]{placeins}

\begin{document}

\title{Federated-Continual Dynamic Segmentation of Histopathology guided by Barlow Continuity}


\author{Niklas Babendererde$^{1}$\\
\and
Haozhe Zhu$^{1}$\\
\and
Moritz Fuchs$^{1}$\\
\and
Jonathan Stieber$^{1}$\\
\and
Anirban Mukhopadhyay$^{1}$\\
\and
{\small$^{1}$ Technical University of Darmstadt},
{\tt\small niklas.babendererde@gris.tu-darmstadt.de}}    

\maketitle

\begin{abstract}
   Federated- and Continual Learning have been established as approaches to enable privacy-aware learning on continuously changing data, as required for deploying AI systems in histopathology images.
However, data shifts can occur in a \textit{dynamic world}, spatially between institutions and temporally, due to changing data over time. This leads to two issues: \textit{Client Drift}, where the central model degrades from aggregating data from clients trained on shifted data, and \textit{Catastrophic Forgetting}, from temporal shifts such as changes in patient populations. Both tend to degrade the model's performance of previously seen data or spatially distributed training.
Despite both problems arising from the same underlying problem of data shifts, existing research addresses them only individually.
In this work, we introduce a method that can jointly alleviate Client Drift and Catastrophic Forgetting by using our proposed Dynamic Barlow Continuity that evaluates client updates on a public reference dataset and uses this to guide the training process to a spatially and temporally shift-invariant model. We evaluate our approach on the histopathology datasets BCSS and Semicol and prove our method to be highly effective by jointly improving the dice score as much as from $15.8\%$ to $71.6\%$ in Client Drift and from $42.5\%$ to $62.8\%$ in Catastrophic Forgetting. This enables \textit{Dynamic Learning} by establishing \textit{spatio-temporal shift-invariance}.
\end{abstract}

\section{Introduction}
\begin{figure*}[]
    \centering
    \includegraphics[width=1.0\textwidth]{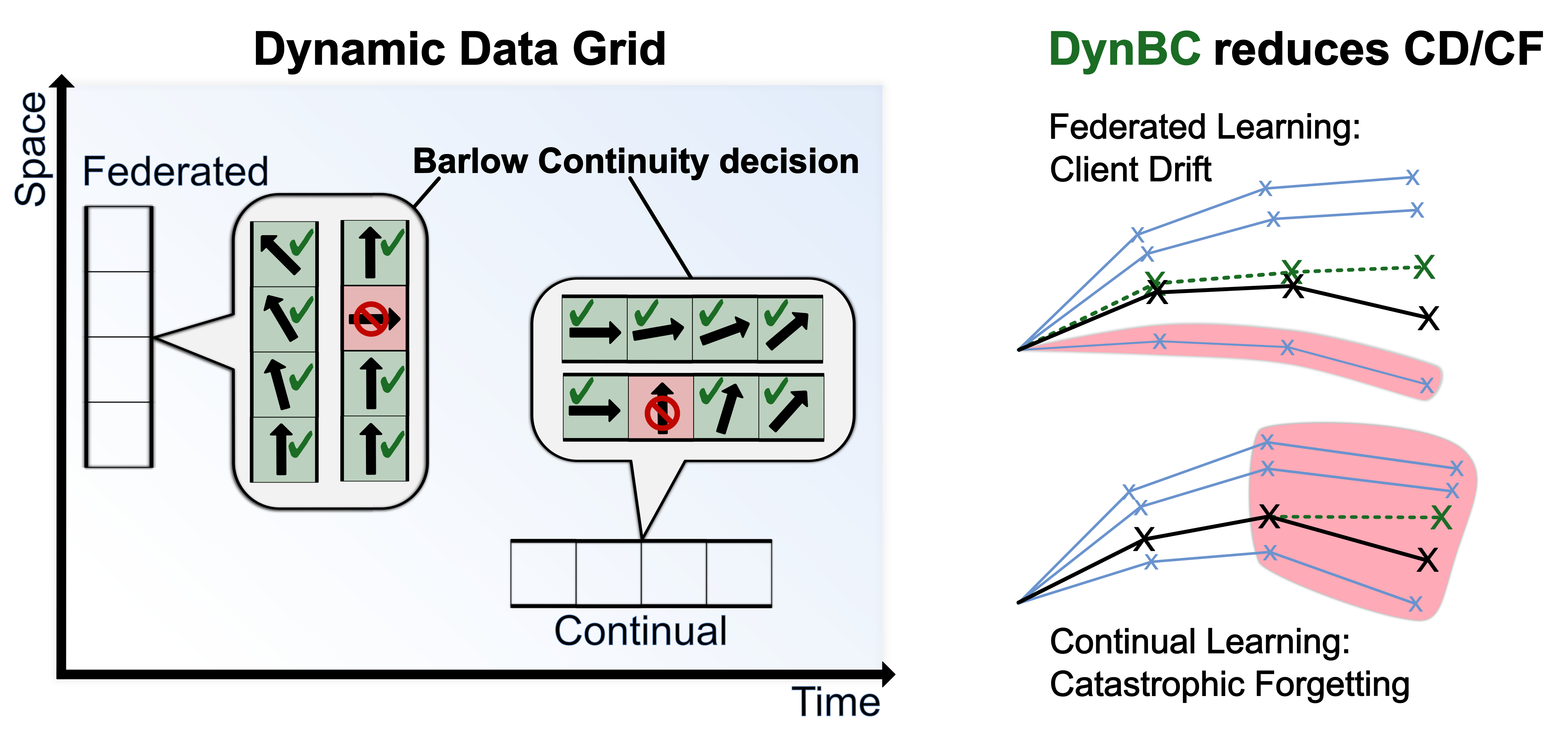}
    \caption{DynBC detects drastic changes of predictions from spatio-temporally trained models (left). It guides the training process to a more robust parameter representation by cancelling such model updates in spatio-temporal scenarios: In Federated Learning (spatial), it cancels model updates from certain clients, while in temporal Continual Learning (temporal) it rolls back to the previous model state, mitigating the problems through distribution shifts (red) such as Client Drift and Catastrophic Forgetting (right).}
    \label{fig:motivation}
\end{figure*}

AI-Assisted histopathology segmentation has the potential to significantly support pathologists and improve experience for their patients.
Generalization of such models \textbf{depends on the availability of diverse data from multiple hospitals}~\cite{bandi2018detection,stacke2020measuring}.
Privacy and logistical requirements typically forbid the sharing of such sensitive patient data. \textbf{Federated Learning (FL)}~\cite{rieke2020future}, which allows decentralized training without sharing of training data, has been established as an important paradigm for circumnavigating the data availability issues of histopathology~\cite{lu2022federated,wagner2022federated,yeom2024comparative}.
On the other hand, privacy and storage requirements necessitate that the model is continuously trained on an evolving data stream without access to the old data~\cite{perkonigg2021dynamic}. This means retraining the model on the entire dataset whenever new data becomes available is impossible. Consequently, the concept of \textbf{Continual Learning}~\cite{lee2020clinical} is also essential for the deployment of privacy-aware AI-assisted histopathology.

However, a core problem in histopathology is that this type of data is prone to \textbf{data shifts}, which manifest spatially between clients in Federated Learning or temporally in Continual Learning~\cite{babendererde2023jointly,bandi2018detection,huo2022comprehensive}.
Spatial shifts, for example, can be caused when certain hospitals use different scanner models for scanning the tissue~\cite{huo2022comprehensive}, staining styles which depend on the preparation technique differ~\cite{stacke2020measuring,wagner2022federated} or artifacts~\cite{stieber2022froodo,taqi2018review} occur in certain hospitals. In Federated Learning, all these types of data shifts can cause performance degradation of the central model, which is known as \textbf{Client Drift (CD)}~\cite{zhao2018federated}.
Temporal Shifts are particularly evident in scenarios such as when new diseases occur~\cite{gonzalez2022distance,murillo_cancer_2022}, the acquisition techniques change~\cite{huo2022comprehensive,wagner2022federated} or the patient population changes~\cite{bandi2018detection,linardos2022federated}.
Continual Learning suffers from \textbf{Catastrophic Forgetting (CF)}\cite{perkonigg2021dynamic} where data shifts over time cause the model to forget old knowledge.
Consequently, addressing both problems of Client Drift and Catastrophic Forgetting is essential for the deployment of AI-assisted histopathology.
However, \textit{existing approaches only address Client Drift and Catastrophic Forgetting individually, despite their underlying common cause of spatio-temporal distribution shifts}.

As medical data is only sparsely available~\cite{lennerz2024dimensions} on the dynamic grid of space-time, assumptions are necessary to make Dynamic Learning feasible. We assume \textbf{spatio-temporal continuity}, meaning \textbf{gradual data drift over space and time corresponds to gradual drift to the main model}~\cite{babendererde2023jointly}. Hence, we conclude that unstable, shifted models are causing drastic prediction changes on minor input variations. Consequently, we evaluate spatio-temporally distributed model updates regarding their continuity.
As a foundation, we leverage the object of Barlow Twins~\cite{zbontar2021barlow}, a widely used method for self-supervised learning~\cite{chen2020simple,grill2020bootstrap} through redundancy reduction.
We adapt it to our goal of evaluating spatio-temporal continuity, introducing the \textbf{Dynamic Barlow Continuity (DynBC)} Objective Function, which employs redundancy reduction to \textbf{detect drastic changes of the model prediction} (Figure~\ref{fig:motivation}).
We apply DynBC during the aggregation step of Dynamic Learning, guiding the global model to a more continuous training process with a parameter representation that is more robust to spatio-temporal shifts. We evaluate our DynBC method in two histopathology datasets -- BCSS~\cite{amgad2019structured} for breast cancer and Semicol~\cite{semicol} for colorectal cancer -- and show significant improvements in both scenarios of Client Drift and Catastrophic Forgetting from a single method. Even though, the application of DynBC is not limited to histopathology, we focus on it in this work as Client Drift and Catastrophic Forgetting are especially prevalent here.
In this work, we introduce the Dynamic Barlow Continuity (DynBC) method, which aids the training process to be more robust to spatio-temporal shifts, allowing Dynamic Learning through joining Federated and Continual Learning. We evaluate this on the histopathology datasets BCSS~\cite{amgad2019structured} and Semicol~\cite{semicol}.

\section{Related Work}
\begin{figure*}[h!]
    \centering
    \includegraphics[width=1.0\textwidth]{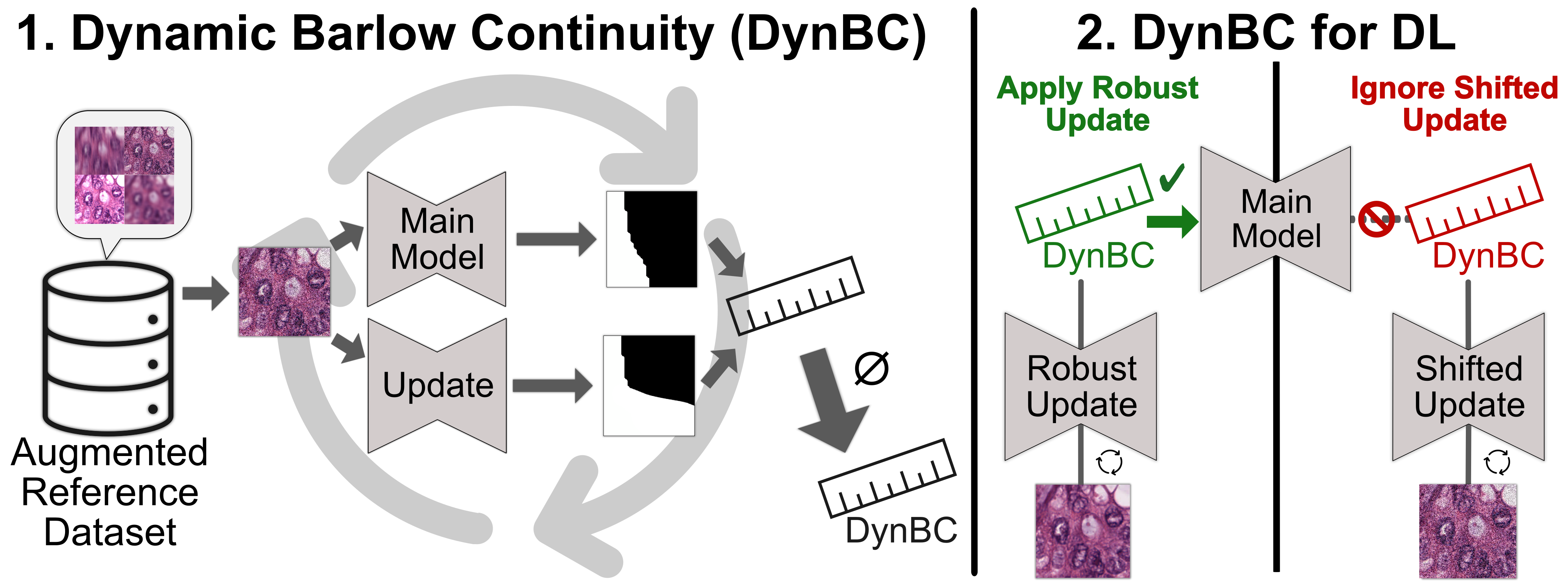}
    \caption{\textbf{DynBC} (1) measures the shift invariance of a model update.
    A small DynBC indicates less biased to a certain distribution, as it shows shift invariance when updating.
    \textbf{DynBC for Dynamic Learning} (2): We compare each potential model update to the previous Continual Learning (CL) or global Federated Learning (FL) model. The resulting distance decides, whether the update is applied or ignored.
    }
    \label{fig:method}
\end{figure*}
In Continual Learning, there are three main branches of learning paradigms: the regularization-based approaches~\cite{kirkpatrick2017overcoming}, the architectural approaches~\cite {aljundi2018memory}, and the replay-based methods~\cite{ranem2024continual,rebuffi2017icarl}. Of these branches, only the replay-based branch has an easy path to be adapted to Federated Learning. The Rehearsal method~\cite{rebuffi2017icarl} is often seen as an upper bound of what can be achieved with such methods. However, it violates patient privacy by saving samples from the old datasets to interleave them during the training of the next stage. Bándi et al.~\cite{bandi2023continual} proposes Continual Learning in the context of histopathology but for task-incremental learning and a classification task, unlike our method that is designed to train segmentation tasks under distribution shifts.

In Federated Learning, many methods that have been proposed can be categorized into either a Weight aggregation approach~\cite{mcmahan2017communication,wagner2022federated,yeom2024comparative,reddiadaptive} or a distillation approach~\cite{hinton2015distilling}. Remarkably, Yoem et al.~\cite{yeom2024comparative} have shown that, in some cases, Federated Learning can even outperform centralized learning in histopathology. Also, recent advances~\cite{bo2023relay} have shown that methods based on learning synthetic datasets and using them to train the server model can be successful they are limited to upper bounds similar to the Rehearsal method~\cite{rebuffi2017icarl} from Continual Learning. 

Federated Continual Learning~\cite{wang2019adaptive} is a paradigm that combines concepts from Federated- and Continual Learning. Most approaches focus on task-incremental learning: FedWeIT~\cite{yoon2021federated} aims to enable task-incremental learning by dividing the parameters into task-specific and global parameters and sharing them between the clients. Asynchronous Federated Continual Learning~\cite{shenaj2023asynchronous} is designed to train multiple clients on the same tasks but in different order. TARGET~\cite{zhang2023target} is a distillation-based approach that transfers knowledge from previous tasks to other clients.
However, most of these approaches only address the goal of task-incremental learning.

To the best of our knowledge, there is no Federated- or Continual Learning method that was designed for a Dynamic Learning setting, having both the spatial- and the temporal domain in mind while enabling robust training under spatio-temporal distribution shifts.

\section{Method}

The intuition of our method is to guide the training process towards a shift-invariant parameter configuration. We achieve this by continuously evaluating model updates both -- spatially for updates from Federated Learning clients and temporally in a Continual Learning setting -- for the curation of all updates during the Federated Continual model aggregation process.
Consequently, our method consists of a component for the evaluation of model updates and an aggregation method that uses the evaluation results for improving spatio-temporal shift-invariance. Algorithm~\ref{algo:method} provides an overview, how we combine the model evaluation and aggregation method. In the following, we will begin with a detailed definition of our model update evaluation function, before explaining, how we apply it during the aggregation process.

\begin{algorithm*}[h]
\caption{Application of DynBC to the dynamic aggregation process to avoid Client Drift and Catastrophic Forgetting}
\label{algo:method}
\begin{algorithmic} 
\STATE $\mathit{th_\delta} \leftarrow \mathit{2.0}$ \algorithmiccomment{Threshold for maximum factor for exceeding the distance $\delta$ between two models (default $2.0$)}
\STATE $\mathit{\delta_{max}} \leftarrow \mathit{0}$ \algorithmiccomment{Current maximum distance $\delta$ between aggregated models}
\STATE $\mathit{model_{server}} \leftarrow \mathit{None}$
\FOR{$\mathit{epoch_{global}}$ in $\mathit{epochs_{global}}$}
    \STATE $\mathit{newmodel_{server}} \leftarrow \mathit{model_{server}}$ 
    \item \algorithmiccomment{Check and filter Federated Learning updates from the clients for spatial continuity}
    \FOR{$\mathit{model_{client}}$ in $\mathit{models_{clients}}$}
        \STATE $\mathit{\delta} \leftarrow \mathit{DynBC(newmodel_{server}, model_{client})}$\algorithmiccomment{Measure distance between client- and global model using DynBC}
        \IF{$\mathit{\delta}>\mathit{\delta_{max}}$}
            \IF{$\mathit{\delta}<=\mathit{th_\delta}*\mathit{\delta_{max}}$}
                \item $\mathit{newmodel_{server}} \leftarrow \mathit{Aggregate(newmodel_{server}, model_{client})}$
                \STATE $\mathit{\delta_{max}} \leftarrow \mathit{\delta}$
            \ENDIF
            \STATE $\mathit{\delta_{max}} \leftarrow \mathit{\delta}$
        \ELSE
                \item $\mathit{newmodel_{server}} \leftarrow \mathit{Aggregate(newmodel_{server}, model_{client})}$
        \ENDIF
    \ENDFOR
    \item \algorithmiccomment{Check and filter Continual Learning updates for temporal continuity}
    \IF{$\mathit{DynBC(model_{server}, newmodel_{server})}\leq\mathit{th_\delta}*\mathit{\delta_{max}}$}
        \item $\mathit{model_{server}} \leftarrow \mathit{newmodel_{server}}$
    \ENDIF
\ENDFOR
\end{algorithmic}
\end{algorithm*}

\subsection{Evaluation of Model Updates}
The foundation of our method is the evaluation method for potential model updates that we apply on the \textbf{server side during aggregation}.
Our model update evaluation is inspired by the  Barlow Twin Objective Function~\cite{zbontar2021barlow} which is originally designed to evaluate self-supervised learning by measuring the changes of the embeddings of \textbf{one specific image when the input data is shifted}. It has been successfully deployed in the conventional centralized training on closed datasets that it was designed for, but for Dynamic Learning, we need to enable decentralized training (Federated Learning) and training on an open dataset (Continual Learning). Consequently, the comparison of two model states -- from Federated Learning aggregation or Continual Learning model updates -- regarding diverse data is required. Therefore, as visualized in Figure~\ref{fig:method} (left), \textbf{our modified DynBC} function \textbf{compares two models} -- a potential update and the current model state -- instead of only one. The novelty of our evaluation function in comparison with the existing objective function of Barlow Twins stems from the approach of comparing two model states instead of only evaluating one, enabling the live-evaluation of spatio-temporal training processes.
Moreover, to preserve privacy during the model evaluation, we sample reference data from a separate, public dataset of histopathology images. We augment it with one out of a selection of augmentations that changes by each image sample, allowing us to measure the reaction of the model to distribution shifts. We feed the resulting augmented patches into both models that we want to compare. Then we compute the norm between both resulting segmentations. We repeat this step multiple times while sampling different histopathology patches and augmentations. The resulting average of these norms between the segmentation predictions is the \textbf{DynBC}. We pass each of the sample patches $x\in D_x$ through the current server model $m_s$ and the proposed updated model $m_c$ from the client $c$. We employ a single augmentation $a$ from a set of augmentations on every sample, ensuring an equal distribution of each augmentation across the samples. The DynBC then calculates as follows: 
\begin{equation}
    DynBC=\frac{\sum_{n=0}^{N} \langle m_s(a(x_n)), m_c(a(x_n))\rangle}{N}
\end{equation}
, where $N$ is the number of available patches.
A small distance between two model states means that the update does not cause sudden changes of predictions on the diverse reference data, which is a desirable property of model updates in Dynamic Learning as the clients are more homogenous within our assumption of \textbf{spatio-temporal continuity} as it is \textbf{indicating more shift invariance}. Therefore, DynBC handles a sudden increase in the distance between model updates as an indicator for an update that decreases shift invariance and therefore would make the newly aggregated central model of a Federated Learning setup more prone to Client Drift and an updated model of a Continual Learning Setup experience Catastrophic Forgetting. On the other hand, a limited increase must be tolerated to allow training progress. We use a dynamic threshold to balance both goals. Therefore, after an initialization phase, we measure the maximum distance that was experienced so far. Models, that exceed this maximum distance by more than a given factor, are ignored. We chose the default threshold factor of 2.0 in this work as it showed good performance. However, this parameter still has potential for further fine-tuning to improve the performance further. Using a factor instead of an absolute distance as threshold, ensures model independence in the choice of this DynBC hyperparameter.

\subsubsection{Choice of metric for DynBC}
As an essential part of DynBC distance measurement is the comparison of the two segmentation masks derived from the compared model states, we explain in the following the reasoning behind using the dot product as the underlying similarity metric.
A first important criterion specifically for the given histopathology context is that the goal is the precise measurement of overlap of the segmentation. The dot product allows us to directly measure this, unlike other metrics such as Cosine Similarity or KL-Divergence.
Besides the precision, also computational efficiency led to the decision to use the dot product. In the case of histopathology segmentation masks, there are often masks with wide areas annotated with the label zero. The dot product is efficient at operating on such sparse data, as only non-zero values need to be considered.

\subsection{Improving robustness against Client Drift}
We deploy the method at the \textbf{aggregation step} of the server by \textbf{evaluating the update from each Client} to the current main server model. Consequently, we apply or ignore the update from a client based on the DynBC distance. As visualized in Figure~\ref{fig:method} (right), the goal is to detect an uncommon increase in the DynBC for certain client update, as we want to limit the global model update to only gradual drifts. Such an increase would exceed the aforementioned threshold factor. Therefore, we track the global maximum DynBC, and if we surpass it by a certain factor, this drastic change is considered an indicator of an undesirable update regarding shift-invariance. In such cases, the server ignores the update from the specific client and only averages from clients with sufficiently low DynBC. To ensure the method's reliability, DynBCs that are exceeding their previous maximum by the given factor are not tracked among the maximum DynBCs. The latest global update is distributed to the ignored client to guide it to a more robust parameter state along with the other clients, \textbf{decreasing the performance degradation from Client Drift}.

\subsection{Improving robustness against Catastrophic Forgetting}
Similarly to the Client Drift scenario, we use our method to evaluate model updates in Continual Learning regarding their shift-invariance and ensure homogeneity in the temporal domain.
Therefore, we \textbf{evaluate each potential model update} and the corresponding current model using our DynBC metric.
We track the maximum DynBC over all valid model updates and ignore updates that surpass it by a given factor. This way, we \textbf{guide the model to a more shift-invariant and homogenous distribution}, helping to \textbf{decrease the performance degradation from Catastrophic Forgetting}.

\section{Experiments and Results}
\begin{figure}[]
    \centering
    \includegraphics[width=\columnwidth]{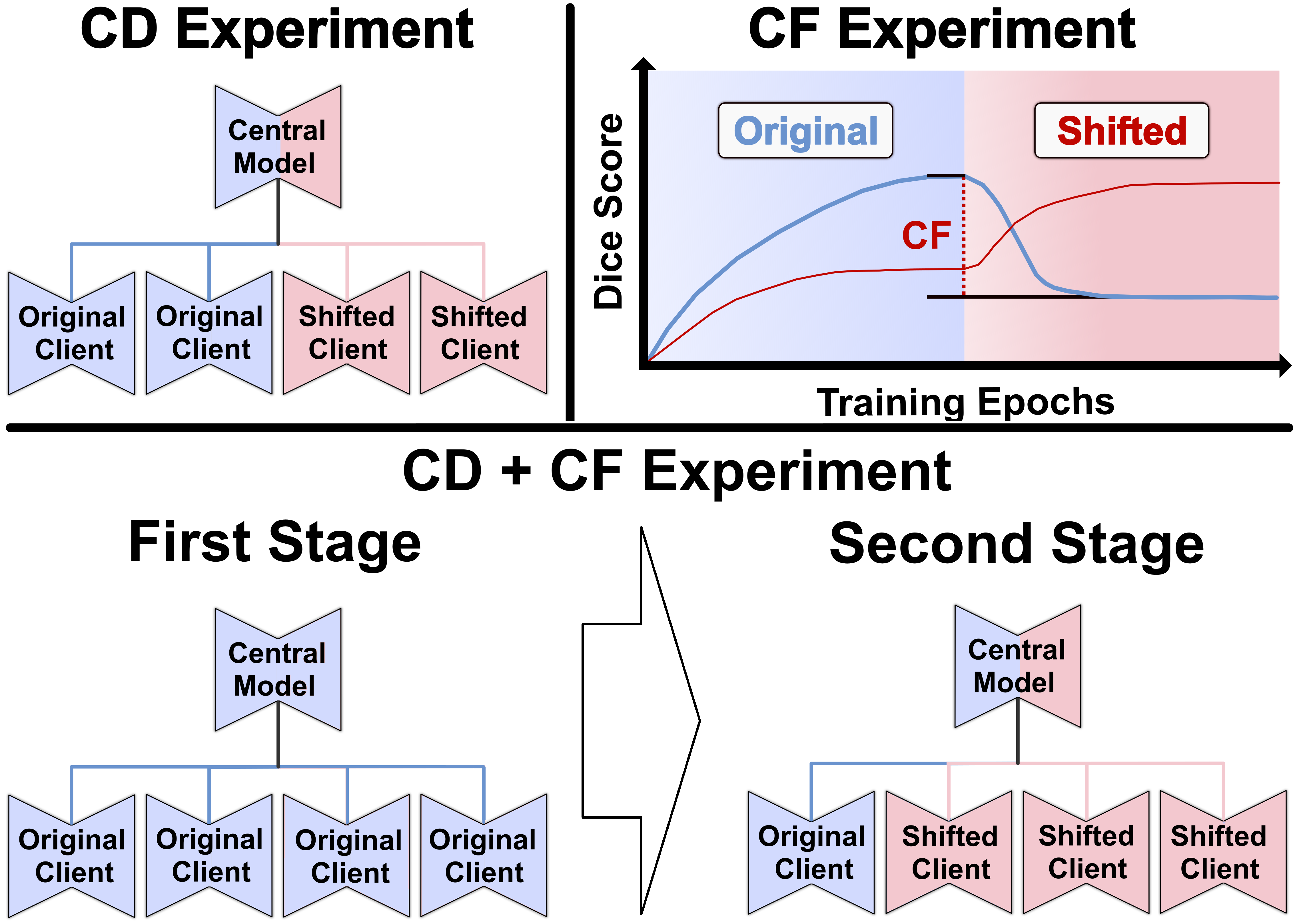}
    \caption{Setting of the experiments for Client Drift, Catastrophic Forgetting and their combination}
    \label{fig:exp}
\end{figure}
In the following section, we provide details on the datasets and setup of the experiments for evaluating DynBC. Then we evaluate DynBC in the scenarios of Client Drift, Catastrophic Forgetting and their combination and compare the resulting performance improvements to our baseline and multiple comparison methods such as the \textit{upper-bound} of Rehearsal from the Continual Learning domain, FedAdam from Federated Learning and FedWeIT from Federated Continual Learning. Lastly, we conduct ablation studies, comparing different threshold parameters and showcasing that DynBC does benefit from augmentations on the reference dataset regarding spatio-temporal shifts.

\subsection{Dataset and Experimental Setup}
\begin{figure*}[]
  \centering
  \begin{floatrow}
    \ffigbox[\FBwidth]
      {\includegraphics[width=\linewidth]{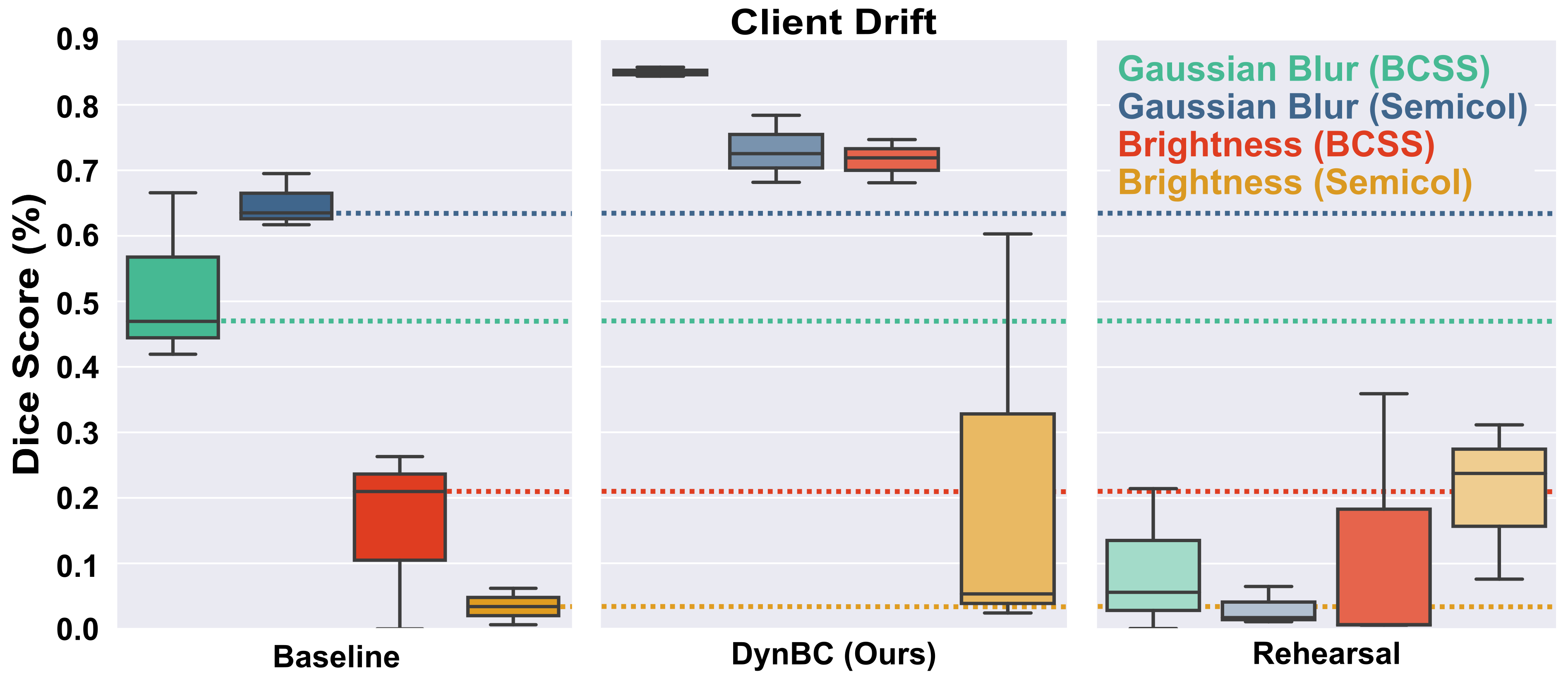}}
      {\caption{Comparison of our method (DynBC) with Rehearsal and the baseline without any method on the datasets BCSS and Semicol in CD}\label{fig:cd}}
    \ffigbox[\FBwidth]
      {\includegraphics[width=\linewidth]{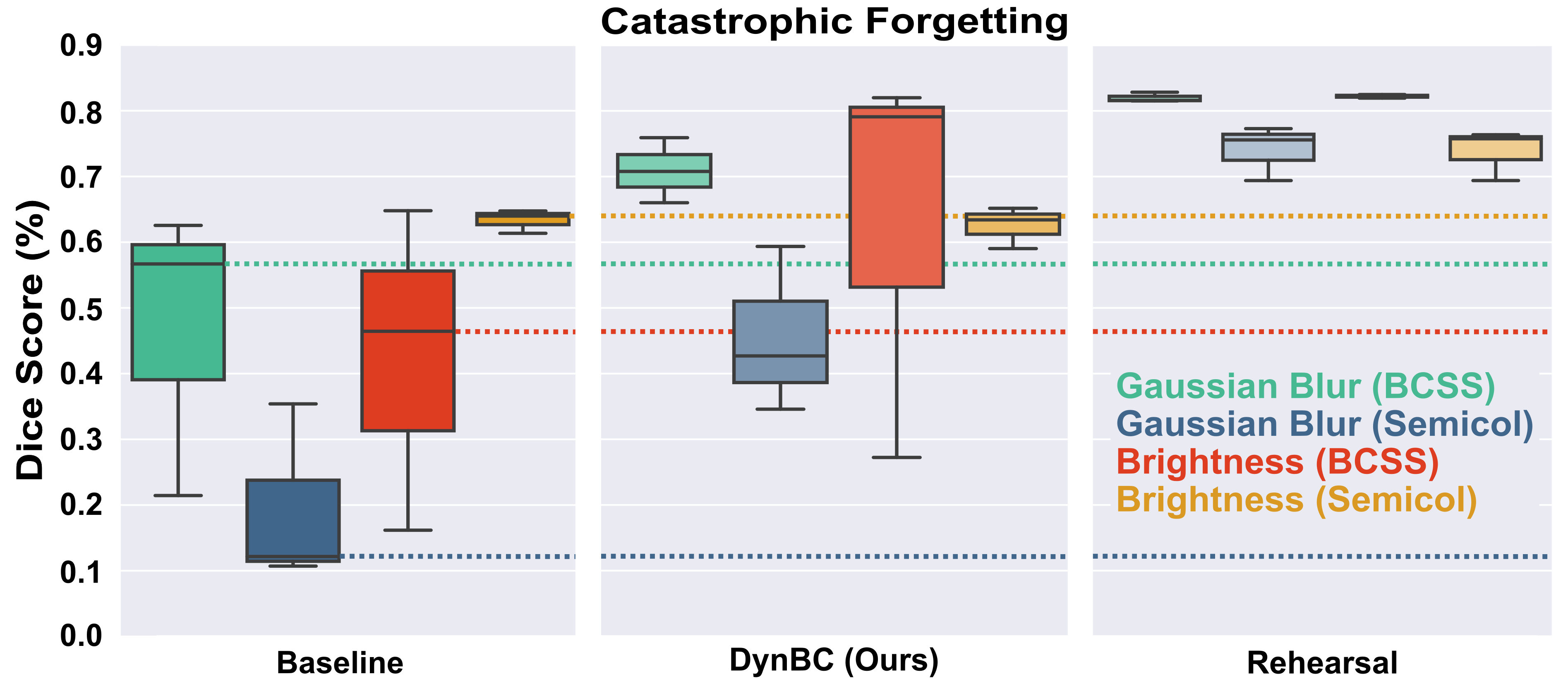}}
      {\caption{Comparison of our method (DynBC) with Rehearsal and the baseline without any method on the datasets BCSS and Semicol in CF}\label{fig:cf}}
  \end{floatrow}
\end{figure*}

We perform supervised training of a U-Net~\cite{ronneberger2015u}, using \emph{BCEWithLogitsLoss} from PyTorch~\cite{paszke2019pytorch} with ADAM~\cite{kingma2014adam} as optimizer. The learning rate is $1e^{-4}$ and the batch size is $4$. We use three public histopathology segmentation datasets: BCSS~\cite{amgad2019structured} and Semicol~\cite{semicol} for the evaluation and Camelyon17~\cite{bandi2018detection} as a separate reference dataset for DynBC. All datasets use a patch size of $256\times256$ pixel, with 0.25 microns per pixel, and are split by patients. BCSS we split based on patients into $60\%$ train, $30\%$ test, and $10\%$ validation data from $10710$ patches. For Semicol, we sample $70\%$ train, $20\%$ test, and $10\%$ validation data from $2120$ patches. We use Camelyon17 only for measuring prediction distances in DynBC and not for training or testing, and we only sample from $1125$ randomly chosen validation patches. We train a binary segmentation task. Therefore, in BCSS we segment for Tumor and in Semicol for a merged label Tumor and Tumor Stroma. As reference images need to be augmented for DynBC, we augment each sample with one selected augmentation: From \emph{Torchvision}~\cite{marcel2010torchvision} Gaussian Blur with kernel size 19 and $\sigma=4.0$, Motion Blur with a blur limit of 29 or from \emph{Albumentations}~\cite{buslaev2020albumentations} Gaussian Noise with a variance limit of 1000.
We evaluate DynBC (with a maximum factor for DynBC increase of $2$) separately on the public histopathology segmentation datasets BCSS and Semicol and for emulating data shift on the training data, we either apply Torchvision brightness (increase by $2.0$ for BCSS or $1.2$ for Semicol) or Torchvision Gaussian blur with a kernel size of 19 and $\sigma=8.0$.
To avoid data leakage, we don't use Gaussian blur on the DynBC reference dataset when we evaluate this augmentation.
We repeat all experiments on 3 seeds and train on a GeForce RTX 4090 and A40 GPUs using CUDA SDK 11.6 and PyTorch 1.13.1.

\subsection{Evaluation for Client Drift}

\begin{table*}[]
\begin{tabular}{|c|cc|cc|}
\hline
\textbf{Dataset}    & \multicolumn{2}{c|}{BCSS}                                          & \multicolumn{2}{c|}{Semicol}                                     \\ \hline
\textbf{Shift Type} & \multicolumn{1}{c|}{Gaussian Blur}         & Brightness            & \multicolumn{1}{c|}{Gaussian Blur}       & Brightness            \\ \hline
Baseline            & \multicolumn{1}{c|}{$0.518\pm 0.13$}           & $0.158\pm 0.139$          & \multicolumn{1}{c|}{$0.649\pm 0.04$}         & $0.034\pm 0.028$          \\ \hline
DynBC (Ours)        & \multicolumn{1}{c|}{$\mathbf{0.849\pm 0.007}$} & $\mathbf{0.715\pm 0.033}$ & \multicolumn{1}{c|}{$\mathbf{0.73\pm 0.05}$} & $\mathbf{0.227\pm 0.326}$ \\ \hline
Rehearsal           & \multicolumn{1}{c|}{$0.09\pm 0.11$}            & $0.124\pm 0.203$          & \multicolumn{1}{c|}{$0.031\pm 0.029$}        &$0.209\pm 0.12$           \\ \hline
\end{tabular}
\caption{Evaluation in Client Drift scenario with mean and standard deviation of dice scores}
\label{tab:cd}
\end{table*}
In the case of BCSS, we deploy 5 clients while 3 clients experience a data shift via augmentations and on Semicol, we run with 4 clients, and 2 of them augmented (Figure \ref{fig:exp} on the left side).
All clients have their equally sized and individual parts of the train set.
We use FedAvg~\cite{mcmahan2017communication} as algorithm for Federated Learning and train until the global model converges after 100 epochs (BCSS) or 80 epochs(Semicol). We evaluate the central model on a separate original test without augmentations. To ensure consistent results, we train for 20 more epochs and average the test set dice score.
Semicol converges after epoch 80. Similarly, we compute the average dice score over the next 20 epochs.

For comparison, we plot the resulting dice scores of the baseline evaluation without any method against Client Drift or Catastrophic Forgetting, our method DynBC, and Rehearsal~\cite{rebuffi2017icarl} with $10\%$ Rehearsal data. Rehearsal is a well-established method from Continual Learning but can also be applied in a Dynamic scenario~\cite{babendererde2023jointly}. We visualize this comparison in Figure~\ref{fig:cd} and Table~\ref{tab:cd} shows the corresponding dice scores.
The results show a consistent improvement throughout all scenarios. In Client Drift on BCSS and Semicol (Gaussian blur), DynBC is always performing significantly better than Rehearsal and the baseline. Only in the case of the brightness-augmented Semicol, the improvement through DynBC is inconsistent. We explain this with information loss through the brightness augmentation, as the patches from Semicol are already bright. Rehearsal can compensate for this by training partially on the original data distribution. However, it is not privacy-preserving. Therefore, the improvements from Rehearsal come at the price of breaking the assumption that there is no access to the original distribution.
In all other experiments for Client Drift, Rehearsal even shows a performance degradation compared to the baseline, as it is originally a method from Continual Learning and can't mitigate the spatial shift. DynBC benefits from its more holistic approach that generally improves shift invariance by aiming for spatio-temporal continuity.

\subsection{Evaluation for Catastrophic Forgetting}
\begin{table*}[]
\begin{tabular}{|c|cc|cc|}
\hline
\textbf{Dataset}    & \multicolumn{2}{c|}{BCSS}                                          & \multicolumn{2}{c|}{Semicol}                                     \\ \hline
\textbf{Shift Type} & \multicolumn{1}{c|}{Gaussian Blur}         & Brightness            & \multicolumn{1}{c|}{Gaussian Blur}       & Brightness            \\ \hline
Baseline            & \multicolumn{1}{c|}{$0.469\pm 0.222$}           & $0.425\pm 0.245$          & \multicolumn{1}{c|}{$0.194\pm 0.138$}         & $0.634\pm 0.018$          \\ \hline
DynBC (Ours)        & \multicolumn{1}{c|}{$0.709\pm 0.049$} & $0.628\pm 0.308$ & \multicolumn{1}{c|}{$0.455\pm 0.126$} & $0.625\pm 0.031$ \\ \hline
Rehearsal           & \multicolumn{1}{c|}{$\mathbf{0.82\pm 0.007}$}            & $\mathbf{0.822\pm 0.002}$          & \multicolumn{1}{c|}{$\mathbf{0.741\pm 0.042}$}        &$\mathbf{0.738\pm 0.039}$           \\ \hline
\end{tabular}
\caption{Evaluation in Catastrophic Forgetting scenario with mean and standard deviation of dice scores}
\label{tab:cf}
\end{table*}
As in the evaluation for Client Drift, we emulate data shift by applying the same selection of augmentations on the same two histopathology datasets.
However, we want to simulate the performance impact of a train set that shifts over time when using our method compared to the baseline and Rehearsal with $10\%$ rehearsed data.
First, we train a single model on the original train set without applying any shift for 75 epochs before switching to one of the selected augmentations to emulate data shift and continue training on the shifted data until convergence (130 BCSS and 80 Semicol).
Catastrophic Forgetting shows a performance drop after shifting to the augmented dataset (Figure \ref{fig:exp} on the right side).
For consistent results, we again evaluate the average test set dice score of the last 20 epochs after convergence.
According to Figure~\ref{fig:cf} and Table~\ref{tab:cf}, DynBC significantly improves the performance in all scenarios except the brightness-augmented Semicol. However, the performance loss is minimal ($0.6336$ to $0.6253$ dice score) compared to the improvements in all other scenarios.
Rehearsal gives the best performance in this scenario, but this is not surprising, as it is often seen as an upper bound for preserving performance in scenarios of Catastrophic Forgetting because it comes at the price of breaking the assumption that old data cannot be accessed anymore. In privacy-sensitive applications such as training on histopathology images, this violation would often be problematic due to privacy regulations. DynBC is preserving this principle as it solely relies on a separate dataset, which can also be a public dataset, as in this evaluation. 

\subsection{Evaluation for combined Client Drift and Catastrophic Forgetting}
\begin{table*}[]
\begin{tabular}{|c|cc|cc|}
\hline
\textbf{Dataset}    & \multicolumn{2}{c|}{BCSS}                         & \multicolumn{2}{c|}{Semicol}                      \\ \hline
\textbf{Shift Type} & \multicolumn{1}{c|}{Gaussian Blur} & Brightness   & \multicolumn{1}{c|}{Gaussian Blur} & Brightness   \\ \hline
Baseline            & \multicolumn{1}{c|}{$0.579\pm0.062$}  & $0.479\pm0.151$ & \multicolumn{1}{c|}{$0.077\pm0.058$}  & $0.2\pm0.129$   \\ \hline
DynBC (Ours)        & \multicolumn{1}{c|}{$\mathbf{0.698\pm0.21}$}   & $0.731\pm0.118$ & \multicolumn{1}{c|}{$\mathbf{0.518\pm0.226}$}  & $\mathbf{0.733\pm0.031}$ \\ \hline
Rehearsal           & \multicolumn{1}{c|}{$0.669\pm0.106$}  & $0.682\pm0.1$   & \multicolumn{1}{c|}{$0.353\pm0.231$}  & $0.328\pm0.212$ \\ \hline
FedAdam             & \multicolumn{1}{c|}{$0.553\pm0.479$}  & $0.553\pm0.479$ & \multicolumn{1}{c|}{$0.014\pm0.024$}  & $0.008\pm0.012$ \\ \hhline{|=|=|=|=|=|}
FedWeIT$^*$         & \multicolumn{1}{c|}{-}          & $\mathbf{0.758\pm0.102}$ & \multicolumn{1}{c|}{-}          & $0.728\pm0.043$         \\ \hline
\multicolumn{5}{|l|}{$^*$Upper bound task-incremental comparison method that requires distribution information} \\ \hline
\end{tabular}
\caption{Evaluation in combined Client Drift and Catastrophic Forgetting scenario with mean and standard deviation of dice scores}
\label{tab:fcl}
\end{table*}

Our method is designed to concurrently mitigate Client Drift and Catastrophic Forgetting in a setting of Federated-Continual Learning. To simulate such a scenario, we split the federated training process into two stages as shown in Figure~\ref{fig:exp}: In the first stage, all 4 clients train on the same histopathology dataset (BCSS or Semicol) without any augmentation for 75 epochs. In the second stage, 3 of the 4 clients are changing to augmented training data for an additional 150 epochs. All shifted clients are sharing the same augmentation type chosen from the augmentation set stated in the pure Client Drift and Catastrophic Forgetting scenarios. We run the experiments on three seeds. To analyze the performance impact from combined Client Drift and Catastrophic Forgetting, we evaluate the central model of the server on the corresponding unshifted test set using the dice score. We compare with the baseline without any method against CD/CF, Rehearsal as method against Catastrophic Forgetting, FedAdam against Client Drift and FedWeIT for Federated Continual Learning. Due to the high computational effort of the multi-stage training process on multiple clients and seeds, we limit the experiments on FedWeIT to the brightness augmentation while still covering both datasets.
Table~\ref{tab:fcl} shows the results of this evaluation. Our method DynBC improves the segmentation performance across all datasets and augmentations. It always outperforms its individual comparison methods against Client Drift (FedAdam) and Catastrophic Forgetting (Rehearsal). \textbf{This highlights the advantage of our fundamental concept to concurrently address both problems compared to existing individual approaches}. Regarding FedAdam, the similar results in case of BCSS between both augmentation types might require for an explanation: FedAdam is an optimization-based method that can potentially optimize the model statically to a certain state. As we are evaluating on clean clinical data, this can lead to identical results when applying FedAdam.

We evaluate FedWeIT as an additional SOTA comparison method for Federated Continual Learning. As a method for task-incremental learning, it requires manually setting task parameters in the beginning, unlike our method that adapts itself due to analyzing the continuity during its whole operation. This means that for this evaluation, we had to set the task-specific parameters according to the augmentation status of the clients in order for it to operate. \textbf{This gives FedWeIT knowledge about the underlying distribution, which our method doesn’t require}.  It makes the deployment of FedWeIT to unknown distribution shifts, such as in most real-world histopathology applications, impossible. Due to the lack of alternative publications that address this problem of spatio-temporal distribution shifts, we decided to include FedWeIT as upper bound SOTA comparison method.
Even under these more favorable conditions for FedWeIT, \textbf{DynBC achieves segmentation performance comparable to this state-of-the-art method}. While FedWeIT marginally surpasses DynBC on BCSS, DynBC demonstrates slightly better performance on Semicol.
DynBC is easier to deploy as it only requires modifications at the aggregation of the server. Furthermore, it \textbf{does not require any prior knowledge of the underlying data distribution, a critical factor for most real-world histopathology applications, highlighting the potential of DynBC}.

\subsection{Ablations}
In the following, we compare different ablations of the DynBC method. We start with analyzing the impact of augmenting the reference dataset and continue with comparing different DynBC threshold factors.
\subsubsection{Augmentation of the Reference Dataset}
We conduct an ablation on whether augmenting the reference dataset improves the performance of DynBC. Table~\ref{tab:ablation} shows the performance on BCSS with brightness augmentation and the same settings as for the experiments on Client Drift and Catastrophic Forgetting. The augmentations improve the diversity of the reference data, which improves the ability of DynBC to evaluate the shift invariance, especially in case of Client Drift. For Catastrophic Forgetting, there is no significant difference, which led us to use the augmentations in both setups.
\begin{table}[]
\centering
\resizebox{\columnwidth}{!}{
\begin{tabular}{|l|l|l|}
\hline
\textbf{Method}          & \textbf{Dice Score (CD)}          & \textbf{Dice Score (CF)}         \\ \hline
\textbf{DynBC with Augmentations}   & $\mathbf{0.716\pm 0.001}$ & $0.628\pm 0.095$ \\ \hline
DynBC without Augmentations & $0.566\pm 0.066$  & $\mathbf{0.655\pm 0.028}$ \\ \hline
\end{tabular}
}
\caption{Ablation of DynBC, evaluated on BCSS with brightness augmentation}
\label{tab:ablation}
\end{table}

\subsubsection{Threshold factor for model updates}
DynBC applies a threshold factor on the measured model distances that determines, whether a model update is considered to fulfill the Continuity requirements and should be applied or if it should be discarded. As the DynBC distance is calculated based on a predicted segmentation mask instead of model parameters, \textbf{the threshold is independent of the underlying model architecture}. However, the choice of a right threshold is fundamental for choosing the right model updates: Too low a threshold would hinder the training process as it would block most of the updates, potentially avoiding any training progress after the initialization phase of DynBC. If the threshold is configured too high, this method would not filter enough unintended model updates and therefore not be able to improve segmentation performance in the setting of Client Drift and Catastrophic Forgetting.
In Table~\ref{tab:ablationth}, we apply DynBC to the Catastrophic Forgetting Scenario on the BCSS dataset with the same brightness- and gaussian blur augmentation as used in the main experiments and evaluate the resulting dice scores.

\begin{table}[]
\resizebox{\columnwidth}{!}{
\begin{tabular}{|c|c|c|}
\hline
\textbf{DynBC Threshold Factor} & \textbf{Brightness} & \textbf{Gaussian Blur} \\ \hline
1.9                             & 0.773               & 0.699                  \\ \hline
\textbf{2.0}                    & \textbf{0.79}       & \textbf{0.708}         \\ \hline
2.1                             & 0.611               & 0.608                  \\ \hline
\end{tabular}
}
\caption{Comparison of dice scores from different DynBC parameters for the threshold factor of the DynBC distance increase to still accept a model update. We evaluate on BCSS augmented with either brightness or gaussian blur in the Catastrophic Forgetting setting}
\label{tab:ablationth}
\end{table}

The results indicate that the threshold factor $2.0$ provides the best performance, supporting our decision to use this parameter in our work.
\subsection{Limitations of DynBC}
Our method requires a separate reference dataset. This limits the application in scenarios with very limited availability of public training data. As in this paper we aim for a histopathology-based application, this is not a problem as the availability of public training data is generally given in this field, even though these typically cannot cover the large variance of distribution shifts that occur over time or between hospitals. DynBC helps to mitigate these challenges, helping to enable robust deployment of AI for histopathology.

\section{Conclusion}
We introduce \textbf{Dynamic Barlow Continuity (DynBC)} as a method that guides the training process to \textbf{spatio-temporal continuity}, jointly mitigating two fundamental problems -- Client Drift and Catastrophic Forgetting -- for the robust- and privacy-aware deployment of AI-assisted histopathology.
Our evaluation of two histopathology datasets shows significant improvement in dice score for Client Drift and Catastrophic Forgetting even when they are combined. This demonstrates the \textbf{spatio-temporal shift-invariance for Dynamic Learning} through DynBC.

\section{Acknowledgements}
This work has been partially funded by the Federal Ministry of Education and Research as part of the Software Campus project "FedVS4Hist" (grant 01IS23067) and as part of the project "FED-PATH" (grant 01KD2210B).

{\small
\bibliographystyle{ieee_fullname}
\bibliography{egbib}
}

\end{document}